\documentclass{article}
\usepackage{amsmath,epsfig}
\usepackage[preprint]{spconfa4}
\usepackage{multirow}
\usepackage{color}
\usepackage{fancyhdr}

\copyrightnotice{Copyright notice – 978-1-7281-1331-9/20/\$31.00 ©2020 IEEE}

\let\OLDthebibliography\thebibliography
\renewcommand\thebibliography[1]{
  \OLDthebibliography{#1}
  \setlength{\parskip}{0pt}
  \setlength{\itemsep}{0pt plus 0.3ex}
}

\begin{document}\sloppy

\def\x{{\mathbf x}}
\def\L{{\cal L}}
\def\lyu{\textcolor{black}}

\title{MutAtt: visual-textual mutual guidance for Referring Expression Comprehension}
%
\name{Shuai Wang$^{\ast}$, Fan Lyu$^{\ast}$ ,Wei Feng$^{\ast}$ ,Song Wang$^{\ast}$$^{\dagger}$}
\address{
	$^{\ast}$College of Intelligence and Computing, Tianjin University, China\\
	$^{\dagger}$Department of Computer Science and Engineering, University of South Carolina, US \\
	\{wangshuai201909, fanlyu\}@tju.edu.cn, wfeng@ieee.org, songwang@cec.sc.edu}

\maketitle

\begin{abstract}
Referring expression comprehension (REC) aims to localize a text-related region in a given image by a referring expression in natural language. 
Existing methods focus on how to build convincing visual and language representations independently, which may significantly isolate visual and language information. In this paper, we argue that for REC the referring expression and the target region are semantically correlated and subject, location and relationship consistency exist between vision and language.
On top of this, we propose a novel approach called MutAtt to construct mutual guidance between vision and language, which treat vision and language equally thus yield compact information matching. Specifically, for each module of subject, location and relationship, MutAtt builds two kinds of attention-based mutual guidance strategies. One strategy is to generate vision-guided language embedding for the sake of matching relevant visual feature. The other reversely generates language-guided visual feature to match relevant language embedding. This mutual guidance strategy can effectively guarantees the vision-language consistency in three modules. Experiments on three popular REC datasets demonstrate that the proposed approach outperforms the current state-of-the-art methods.
\end{abstract}
\begin{keywords}
	Referring expression comprehension,  vision-language matching, mutual guidance
\end{keywords}

\renewcommand{\thefootnote}{}
\footnote{\noindent This work was supported by the National Natural Science Foundation of China (Nos. U1803264, 61672376, 61671325).}

\section{Introduction}
Referring expression comprehension (REC), also known as visual grounding, aims at finding the text-related object in a given image according to the description of referring expressions. 
\lyu{As a vision-language problem, REC has widespread applications in real-world scenarios}, e.g., in an autopilot system, we need to localize the exact location in images or videos from text expressions like ``park the car on the right side''.
Although much progress has been made in REC, grounding referring expressions remains challenging because it requires a comprehensive understanding of complex language semantics and various types of visual information simultaneously.


\begin{figure}[t]
	\centering
	\includegraphics[width=\linewidth]{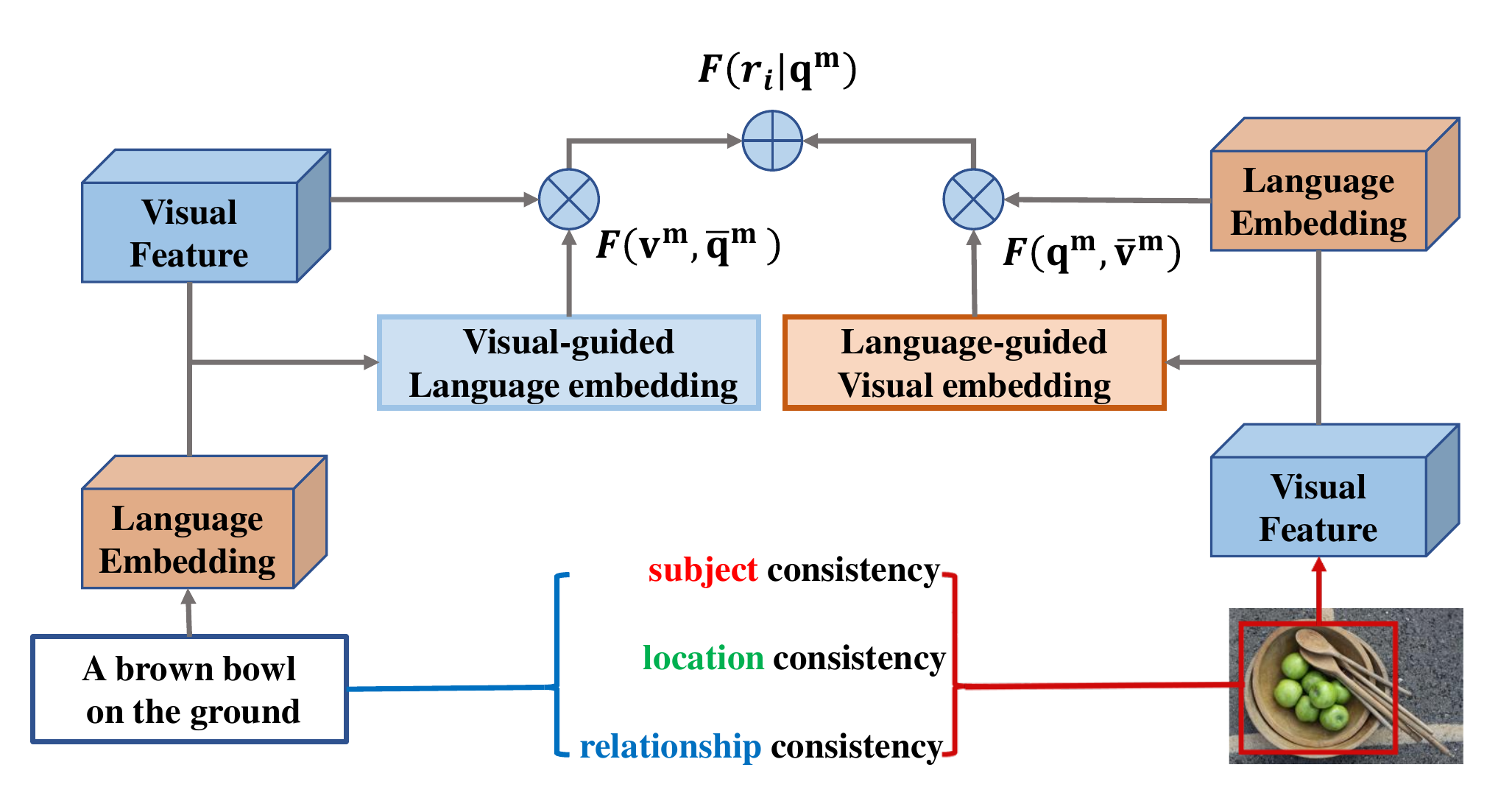} 
	\caption{
		Schematic of the proposed MutAtt.
		We assume there exist three kinds of consistency between referring expression and target region proposal.
		MutAtt builds mutual attention-based guidance strategy between visual and language information, which consists of visual-guided language embedding and language-guided visual embedding.
	}
	\label{fig:1}
	\vspace{-19px}
\end{figure}

\lyu{Researches on REC can be categorized into generative methods and discriminative methods. 
Generative methods, originated from image captioning~\cite{gu2018stack,yang2019learning}, generate description for each localized region in searching by maximum posteriori probability~\cite{yu2017joint,luo2017comprehension,mao2016generation}.
However, generative methods over-relies on the local region captioning model, which cannot describe the relative location and relationships with other objects.
Discriminative methods try to learn the joint vision-language matching score and select object by ranking all scores~\cite{yu2016modeling,hu2017modeling,yu2018mattnet,zhang2018grounding}, which has become the most common ways in REC.}
Existing discriminative methods always focus on how to extract more powerful visual and language features.
Generally, these methods use Convolutional Neural Networks to encode the visual features for each candidate region, and use Recurrent Neural Networks to encode the referring expression~\cite{yu2016modeling,yu2018rethinking}.  
Compositional modular networks~\cite{hu2017modeling,yu2018mattnet} decompose the referring expression into three parts: subject, location and relationship, and design three visual feature representations to achieve fine-grained matching. 
Variational context~\cite{zhang2018grounding} exploits the reciprocal relation between the referent and context to solve the problem of complex context modeling in referring expression comprehension.
\lyu{
Nevertheless, the previous discriminative methods focus on how to build convincing visual and language representations independently, where the referring expressions are always only treated as unrequited queries.
This may significantly isolate visual and language information, thus hinders the effective matching between vision and language, especially when the scene or expression are complex. 
}

\lyu{In our view, \textit{REC can only work based on the hypothesis that the referring expression and the target region represent the same semantics}, including \textbf{subject consistency}, \textbf{location consistency} and \textbf{relationship consistency}.
By considering these three kinds of consistency, REC model can achieve more compact vision and language combination and more accurate prediction.
Based on this hypothesis, we design an innovative mutual attention-based guidance method MutAtt in the perspective of vision-language matching by enforcing these three consistencies.
Specifically, to ensure effective cross modal consistency, we first treat REC as a vision-language matching problem in order to make visual and language information equal.}
MutAtt provides two strategies to achieve the above hypothesis as shown in Fig.~\ref{fig:1}. 
One strategy uses visual features to guide the language and then matches the guided language features with visual features. 
While improving the consistency of cross-modal information, it will make the model focus on vision over language. 
The other strategy uses language embedding to guide vision and then match the generated visual features with language embedding. This allows us to balance the status of vision and language information while further improving cross-model consistency. 
We apply this approach to subject, location and relationship modules, which significantly guarantees three kinds of consistency while maintains vision and language equality.
We conduct experiments on three popular REC datasets to verify the advantages of the proposed method, and the experimental results show the superiority of the proposed MutAtt.

\label{sec:intro}

\section{Related Work}
{\bf Referring expression comprehension.} 
Existing REC methods generally fall into two categories: generative model and discriminative model.
In generative model, \cite{yu2017joint,luo2017comprehension,mao2016generation}use the encoder-decoder structure to localize the region that can generate the sentence with maximum posteriori probability. Discriminative model~\cite{hu2017modeling,yu2018mattnet,zhang2018grounding} tends to use various feature vectors to represent the expression and the image region, and then measures the similarity of them to select the region with the highest scores. The previous work \cite{yu2016modeling} separately encodes the entire related expression and the entire image feature , which ignores the complex structures in the language as well in the image. The work in \cite{hu2017modeling,yu2018mattnet}  overcomes this limitation through decomposing the expression into sub-components and computing the vision-language matching scores of each module. The method~\cite{zhang2018grounding} lowers the requirement of joint grounding and reasoning to a holistic association score between the sentence and region features. In addition, recent work~\cite{yu2018mattnet,lyu2019attend} uses the attention mechanism to make the model focus on more critical information and achieves significant effectiveness.
\lyu{
	However, the previous discriminative methods focus on how to build convincing visual and language representations independently, and never consider the information consistency between vision and language.
	These methods only regard referring expression as a complementary query and overemphasize the importance of visual information.
	In contrast, we propose to enhance the vision-language consistency by cross-modal attention-based mutual guided matching.}


\noindent
\textbf{Vision-language matching.} 
\lyu{
Vision-language matching has been studied for years, the key challenge of which is measuring the similarity between vision and language embedding.}
The most popular vision-language matching methods~\cite{jing2018cascade,li2017identity,li2017person} rely on relatively similar procedures: extract discriminative visual and language features and measure as accurately as possible the distance between the two representations. The work in \cite{frome2013devise,kiros2014unifying} adopt CNN and Skip-Gram or LSTM to extract feature representations for cross-modal. Then a ranking loss is used to force the model to get closer to the matched vision-language pair and away from the unmatched pair.
\cite{gu2018look} further improve the learning of cross-view feature embedding by incorporating generative objectives.
Through region relationship reasoning and global semantic reasoning,~\cite{li2019visual} enhance image representation to align with the corresponding text caption better.
In this paper, we treat the fusion of visual and language features as a kind of vision-language matching problem to enhance the vision-language consistency to make vision and language play same important role.
\lyu{In this way}, the proposed MutAtt can discover more discriminative joint visual-textual representation.

\section{Method}

\begin{figure*}[t]
	\centering
	\includegraphics[width=\linewidth]{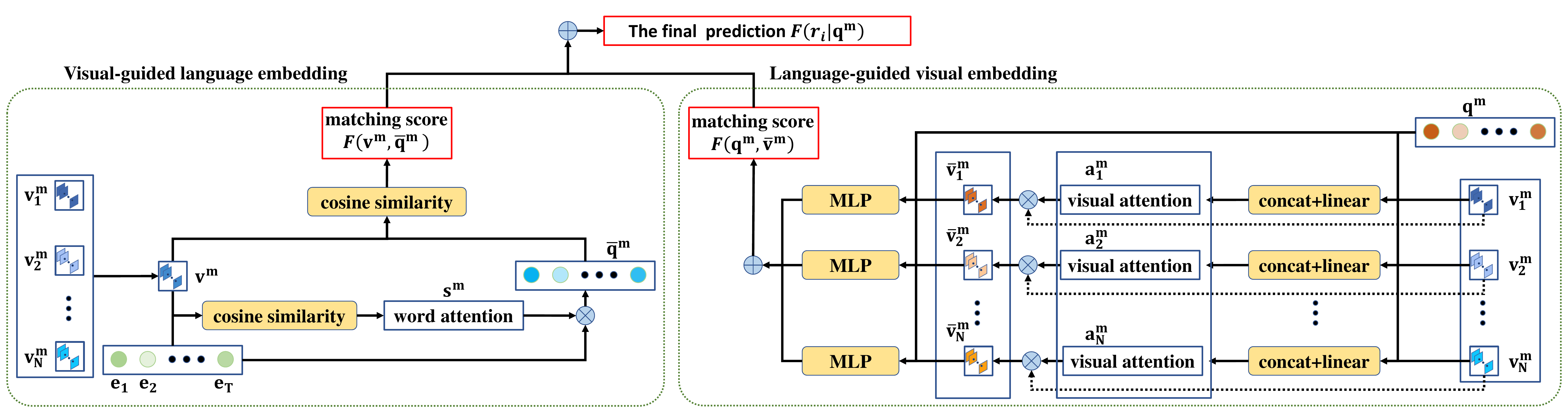} 
	\caption{
		Illustration of MutAtt. 
		$\{\mathbf{v}_{n}^{m}\}_{n=1}^{N}$ represent visual feature of region proposal, $\{\mathbf{e}_{t}\}_{t=1}^{T}$ represent word embedding of sentence and $\mathbf{q}^{m}$ represent phrase embedding of sentence.
		The left part shows visual-guided language embedding, where we compute word attention to guide the generation process of language embedding and match it with visual feature by cosine similarity. The right part shows language-guided visual embedding, where we compute attention on visual feature guided by language embedding and match them by MLPs. Finally, we combine the matching result of two parts as score.
			}
	\label{fig:2}
	\vspace{-15px}
\end{figure*}


\subsection{Problem formulation and background}

Given an image ${I}$ with a set of region of interest $\mathcal{R}=\{r_{i}\}$ tagged by people or detection algorithm and referring expression $\mathcal{E}=\{u_{t}\}_{t=1}^{T}$, where $u_{t}$ means the $t$-th word in sentence, the purpose of REC is to find the target region $r^{*}$ best matching $\mathcal{E}$. The effective solution is to match the visual features of each candidate region and the language embedding of expression, and select the region with the highest score. 
We follow the modular design of MAttNet~\cite{yu2018mattnet} as our backbone for its capability to handle subject, location and relationship information in referring expressions.
MAttNet decomposes expression embedding into three modular components, i.e. $\{\mathbf{q}^\text{subj},\mathbf{q}^\text{loc},\mathbf{q}^\text{rel}\}$, via a language attention network, and designs three visual models to encode the corresponding visual feature $ \mathbf{v}^{m}, $ where $ m \in(\text{subj},\text{loc},\text{rel}) $.
In this paper, we introduce a mutual attention-based guidance approach called MutAtt to improve vision-language consistency, including vision-guided language embedding and language-guided visual feature, which is shown in Fig.~\ref{fig:2}.
\lyu{As we treat the REC problem as a matching problem, we only consider one region $r$ (not specific) in $\mathcal{R}$ as the visual input, while in the inference, the region with largest matching score will be selected.}

\subsection{Mutual attention-based guidance }
\label{method:mutual}
\subsubsection{Visual-guided language embedding}



We first design to use visual feature help the formation of language embedding through matching vision and language from \textit{word-level} to \textit{sentence-level} for each module $m\in \{\text{subj},\text{loc},\text{rel}\}$. 
To be specific, we compute the cosine similarity vector $\mathbf{s}^m$ between word embedding $\{\mathbf{e}_{t}\}_{t=1}^{T}$ and visual feature $\{\mathbf{v}_{n}^{m}\}_{n=1}^{N}$ of region proposal $r$, which can be computed as
\begin{equation}
\mathbf{s}_{t}^{m} = \frac{(\mathbf{v}^{m})^\top\mathbf{e}_{t}}{\left \| \mathbf{v}^{m} \right \|\left \| \mathbf{e}_{t} \right \|},  t\in\left [ 1,T \right ],
\label{eq:sm}
\end{equation}
where $\mathbf{v}^{m}$ is the average pooled visual feature of $\{\mathbf{v}_{n}^{m}\}_{n=1}^{N}$ and can be obtained by 
\begin{equation}
\mathbf{v}^{m} = \frac{1}{N}\sum_{n=1}^{N}\mathbf{v}_{n}^{m},
\end{equation}
where $N$ represents the number of visual element in different module for the candidate region. 
In Eq.~\eqref{eq:sm}, \lyu{$\mathbf{s}_{t}^{m}$} represents the attention from visual feature of module $m$ to the $t$-th word embedding.
By this word-level similarity, we compute the fine-grained similarity between each visual and language element pair, which can significantly compose of the visually-guided language embedding.
Thus, we use the similarity as the weight of each word embedding to generate visual-guided sentence-level embedding as follows:
\lyu{
	\begin{equation}
	\bar{\mathbf{q}}^{m} = \sum_{t=1}^{T} \text{softmax}(\lambda^{m} \mathbf{s}_{t}^{m})\cdot \mathbf{e}_{t},
	\end{equation}
}
where $\lambda^{m}$ is the word-level language attention obtained from language attention network in MAttNet~\cite{yu2018mattnet}, which helps form the language embedding corresponding to different visual modules.
Under the guidance of word-level vision-language similarities, the sentence-level embedding can be enhanced by visual feature. 

After that, we further calculate the score of visual feature $\mathbf{v}^{m}$ and visual-guided language embedding $\bar{\mathbf{q}}^{m}$ by the cosine similarity through matching vision and language in sentence level:
\begin{equation}
	F\left ( \mathbf{v}^{m} , \bar{\mathbf{q}}^{m} \right )=\frac{(\mathbf{v}^{m})^\top\bar{\mathbf{q}}^{m}}{\left \| \mathbf{v}^{m} \right \|\left \| \bar{\mathbf{q}}^{m} \right \|}.
\end{equation}
Note, we propose to match vision and language information from word-level to sentence-level, which can guarantee the multi-scale vision-language matching.
If the region and referring expression never match, the score would be small by this two level matching method, which could help omit failed prediction.
To ensure that vision and language information have equal importance in the matching process and further improve vision-language consistency, we also construct a language-guided visual embedding.


\subsubsection{Language-guided visual embedding}
In our framework, we assume that the language and vision play equal role.
Thus, after using visual information guide language embedding, we also hope to build the reverse guided embedding, i.e., the language-guided visual embedding.
Given the visual feature $\{\mathbf{v}_{n}^{m}\}_{n=1}^{N}$ of region proposal $r$ and the corresponding language embedding $\mathbf{q}^{m}$ of referring expression $\mathcal{E}$, we first compute language-guided visual attention on subject, location and relationship modules.
\begin{equation}
\mathbf{h}_{n} = \tanh(\mathbf{W}_{1}^{m}[\mathbf{v}_{n}^{m},\mathbf{q}^{m}]+\mathbf{b}^{m}),
\end{equation}
\begin{equation}
\mathbf{a}_{n}^{m} = \text{softmax}(\mathbf{W}_{2}^{m}\mathbf{h}_{n}),
\end{equation}
where $[\cdot]$ is the concatenation operation, $\mathbf{W}_{1}^{m}, \mathbf{b}^{m}, \mathbf{W}_{2}^{m}$ are model parameters and $\mathbf{a}_{n}^{m}$ represents the attention from language embedding $\mathbf{q}^{m}$ to the $n$-th visual element of region proposal $r$.
After that, we generate more discriminative language-guided visual feature by
\begin{equation}
\bar{\mathbf{v}}^{m} = \sum_{n=1}^{N}\mathbf{a}_{n}^{m}\cdot \mathbf{v}_{n}^{m}
\end{equation}
\begin{equation}
F({\mathbf{q}}^{m} , \bar{\mathbf{v}} ^{m})=
\text{MLP}({\mathbf{q}}^{m} , \bar{\mathbf{v}} ^{m}).
\end{equation}
Finally, we use MLP structure to calculate the score between language-guided visual feature and language embedding. Each MLP is composed of two fully connected layer with ReLU activation, which help transform cross modal information into a common embedding space. With language-guided visual embedding, we guarantee the consistency of visual and language information and avoid the model paying too much attention to one of them.
Note, the language-guided visual embedding is similar to the common attention using in other REC methods, as they only consider the simple visual language fusion.
The drawback of this method is that the language is treated as the complementary query, while ignore that the their information can guide mutually.


\subsection{Matching result and loss function}

We combine the proposed MutAtt in subject, location and relationship modules. The overall matching score for the region proposal and expression is:
\begin{equation}
F(r_{i}|\mathcal{E})=\omega_{\text{subj}}F(r_{i}|\mathbf{q}^{\text{subj}})+\omega_{\text{loc}}F(r_{i}|\mathbf{q}^{\text{loc}})+\omega_{\text{rel}}F(r_{i}|\mathbf{q}^{\text{rel}}),
\end{equation}
\begin{equation}
F(r_{i}| \mathbf{q}^{m}) = F( \mathbf{v}^{m} , \bar{\mathbf{q}}^{m} ) +F({\mathbf{q}}^{m} , \bar{\mathbf{v}} ^{m}), 
\end{equation}
where $(\omega_{\text{subj}}, \omega_{\text{loc}}, \omega_{\text{rel}})$ represent the weights of subject module, location module and relationship module obtained from language attention network in MAttNet.

For positive candidate object and query pair $(\mathcal{R}_{i},\mathcal{E}_{i})$ and negative pairs  $(\mathcal{R}_{i},\mathcal{E}_{j})$, $(\mathcal{R}_{j},\mathcal{E}_{i})$, the ranking loss is minimized during training: 
\begin{equation}
\begin{aligned}
\text{Loss} = \sum_{i}(&[k-F(\mathcal{R}_{i},\mathcal{E}_{i})+F(\mathcal{R}_{i},\mathcal{E}_{j})]_{+}  \\
+ &[k-F(\mathcal{R}_{i},\mathcal{E}_{i})+F(\mathcal{R}_{j},\mathcal{E}_{i})]_{+}),
\end{aligned}
\end{equation}
where $[x]_{+} = \text{max}(x,0)$, and $k$ is the margin for the loss.


\section{Experiments}

\subsection{Dataset and implementation details}

\textbf{Dataset.}
We use three popular datasets for the evaluation, i.e., RefCOCO, RefCOCO+ and RefCOCOg~\cite{mao2016generation,yu2016modeling}. RefCOCO has 50,000 target objects collected from 19,994 images. RefCOCO+ has 49856 target objects collected from 19,992 images. These two datasets are split into four parts of ``train", ``val", ``testA" and ``testB". RefCOCOg includes 49822 target objects from 25799 images, which are split into three parts of ``train", ``val" and ``test".

\noindent
\textbf{Visual feature representation.} We use faster R-CNN with ResNet101 as backbone to extract subject features, location features and relationship features for each region proposal and follow~\cite{yu2018mattnet} to construct modular visual network.
For the subject network, we feed the whole image into faster R-CNN and extract $7\times7$ features maps from last convolutional output of 3rd-stage and last convolutional output of $4$-th stage to represent subject features. 
For the location network, we represent location features of candidate object by encoding position and relative area as 
$\mathbf{l}_{i} = [\frac{x_{tl}}{W}, \frac{y_{tl}}{H}, \frac{x_{br}}{W}, \frac{y_{br}}{H}, \frac{w\cdot h}{W \cdot H}]$, and encoding relative location offsets and relative areas of up-tp-five surrounding same-category objects as
$\delta \mathbf{l}_{ij} = [\frac{[\Delta x_{tl}]_{ij}}{w_{i}}, \frac{[\Delta y_{tl}]_{ij}}{h_{i}}, \frac{[\Delta x_{br}]_{ij}}{w_{i}}, \frac{[\Delta y_{br}]_{ij}}{h_{i}}, \frac{w_{j} \cdot h}{W \cdot H}]$. 
For the relationship network, we first find up-to-five surrounding objects, then extract their average-pooled visual features and encode their relative position offsets and relative areas to represent relationship features of context objects. For the visual features $\{\mathbf{v}_{n}^{m}\}_{n=1}^{N}$ mentioned in Sec.~\ref{method:mutual}, $n = (49, 1, 5)$ when $m = (\text{subj}, \text{loc}, \text{rel})$.

\noindent
\textbf{Training setting.} The training batch size is 15, which means in each training iteration we feed 15 images and the referring expressions associated with these images to the network. Adam is used as the training optimizer, with initial learning rate to be 0.0004, which decays by a factor of 10 every 8000 iterations. We implement MutAtt based on PyTorch.

\noindent
\textbf{Evaluation setting.}
Following previous work~\cite{wang2019neighbourhood,yang2019dynamic}, we take the region proposals from human annotated (gt) and detection methods (det).
For gt, the evaluation requires the region with the highest matching score is the same as the ground-truth region.
For det, the evaluation requires the intersection over union between the region with highest matching score and ground-truth region is greater than 0.5.


\begin{table*}[t]
	\centering
	\caption{Comparison with state-of-the-art REC approaches on ground-truth regions and  automatically detected regions. It can be seen that our method has significantly improved compared with other methods, and is superior to SOTA in most indicators.}
	\label{table:compare-table}
	\resizebox{.8\linewidth}{!}{
	\begin{tabular}{l|c|cccccc|ccc}
		\hline
		\multirow{2}{*}{Method} & \multirow{2}{*}{Box} & \multicolumn{3}{c|}{RefCOCO}               & \multicolumn{3}{c|}{RefCOCO+} & \multicolumn{3}{c}{RefCOCOg} \\ \cline{3-11} 
		&                               & val   & testA & \multicolumn{1}{c|}{testB} & val      & testA    & testB   & val*     & val     & test    \\ \cline{1-11}
		visdif+MMI~\cite{yu2016modeling}     & gt                  & -     & 73.98 & \multicolumn{1}{c|}{76.59} & -        & 59.17    & 55.62   & 64.02    & -       & -       \\
		Speaker/visdif~\cite{yu2016modeling} & gt                  & 76.18 & 74.39 & \multicolumn{1}{c|}{77.30} & 58.64    & 61.29    & 56.24   & 59.40    & -       & -       \\
		S-L-R~\cite{yu2017joint}          & gt                  & 79.56 & 78.65 & \multicolumn{1}{c|}{80.22} & 62.26    & 64.60    & 59.62   & 72.63    & 71.65   & 71.92   \\
		VC~\cite{zhang2018grounding}             & gt                  & -     & 78.98 & \multicolumn{1}{c|}{82.39} & -        & 62.56    & 62.90   & 73.98    & -       & -       \\
		Attr~\cite{liu2017referring}           & gt                  & -     & 78.05 & \multicolumn{1}{c|}{78.07} & -        & 61.47    & 57.22   & 69.83    & -       & -       \\
		Accu-Att~\cite{deng2018visual}       & gt                  & 81.27 & 81.17 & \multicolumn{1}{c|}{80.01} & 65.56    & 68.76    & 60.63   & 73.18    & -       & -       \\
		PLAN~\cite{zhuang2018parallel}           & gt                  & 81.67 & 80.81 & \multicolumn{1}{c|}{81.32} & 64.18    & 66.31    & 61.46   & 69.47    & -       & -       \\
		Multi-hop Film~\cite{strub2018visual} & gt                  & 84.9  & 87.4  & \multicolumn{1}{c|}{83.1}  & 73.8     & \textbf{78.7}     & 65.8    & 71.5     & -       & -       \\
		MattNet~\cite{yu2018mattnet}        & gt                  & 85.65 & 85.26 & \multicolumn{1}{c|}{84.57} & 71.01    & 75.13    & 66.17   & -        & 78.10   & 78.12   \\
		$\text{NMT}_{\text{REE}}$~\cite{liu2019learning}        & gt                  & 85.65 & 85.63 & \multicolumn{1}{c|}{85.08} & 72.84    & 75.74    & 67.62   & 78.03        & 78.57   & 78.21   \\
		LGRANS~\cite{wang2019neighbourhood}   & gt                  & 82.0 & 81.2 & \multicolumn{1}{c|}{84.0} & 66.6    & 67.6    & 65.5   & -        & 75.4   & 74.7   \\
		DGA \cite{yang2019dynamic}          & gt                  & 86.34     & 86.64     & \multicolumn{1}{c|}{84.79}     & 73.56        &78.31        & \textbf{68.15}       & -        & 80.21       & \textbf{80.26}       \\ \hline
		MutAtt    & gt                  & \textbf{86.58} & \textbf{87.20} & \multicolumn{1}{c|}{\textbf{85.38}} & \textbf{73.69}    & 76.30    & 67.74   &  -        & \textbf{80.37}   & 79.24   \\\hline\hline 
		S-L-R~\cite{yu2017joint}          & det                  & 69.48 & 73.71 & \multicolumn{1}{c|}{64.96} & 55.71    & 60.74    & 48.80   & -        & 60.21   & 59.63   \\
		PLAN~\cite{zhuang2018parallel}           & det                  & -     & 75.31 & \multicolumn{1}{c|}{65.52} & -        & 61.34    & 50.86   & 58.03    & -       & -       \\
		MattNet~\cite{yu2018mattnet}        & det                  & 76.40 & 80.43 & \multicolumn{1}{c|}{69.28} & 64.93    & 70.26    & 56.00   & -        & 66.67   & 67.01   \\
		LGRANS~\cite{wang2019neighbourhood}  & det                  & - & 76.6 & \multicolumn{1}{c|}{66.4} & -    & 64.0    & 53.4   & 62.5        & -   & -   \\
		DGA~\cite{yang2019dynamic}           & det                  & -     & 78.42     & \multicolumn{1}{c|}{65.53}     & -        & 69.07        & 51.99       & -        & -       & 63.28       \\ \hline
		MutAtt    & det                  & \textbf{78.35} & \textbf{82.52} & \multicolumn{1}{c|}{\textbf{71.50}} & \textbf{67.90}       & \textbf{72.60}       & \textbf{58.60}      & -       & \textbf{68.67}      & \textbf{69.03}      \\ \hline
	\end{tabular}}
	\vspace{-15px}
\end{table*}



\subsection{Results}

{\bf Comparisons with State-of-The-Art.} We provide a comparison of our method with other SOTA methods in Table.~\ref{table:compare-table}, including the results of using two settings on three datasets. As can be seen, MutAtt shows the advantage of the proposed approach. On the ground-truth setting, MutAtt is significantly better than the previous method on the RefCOCO dataset, and performs similarly to the previous method on the RefCOCO+ and RefCOCOg datasets. On more important detection settings, We use the features of res101-frcn and compare with other methods. MutAtt outperforms the state-of-the-art on various split sets of the three datasets. It demonstrates that MutAtt can ensure the equality of vision and language in matching and improve the vision-language consistency on subject, location and relationship module.


\begin{table}[]
	\centering
	\caption{Ablation studies on RefCOCOg dataset.}
	\label{table:Ablation-study}
	\resizebox{0.95\linewidth}{!}{
	\begin{tabular}{c|l|c|c}
		\hline
		&                 																			& val           & test          \\ \hline
		1 & MutAtt:subj+loc+rel     																		& 77.96         & 77.14         \\
		2 & MutAtt:subj(V$\rightarrow$L)+loc+rel 														& 79.33         & 78.53         \\
		3 & MutAtt:subj(V$\rightarrow$L+L$\rightarrow$V)+loc+rel  									    & 80.00         & 79.34         \\
		4 & MutAtt:subj(V$\rightleftharpoons$L)+loc(V$\rightleftharpoons$L)+rel  							& 80.35         & 79.03         \\
		5 & MutAtt:subj(V$\rightleftharpoons$L)+loc(V$\rightleftharpoons$L)+rel(V$\rightleftharpoons$L)  & 80.37         & 79.24         \\ \hline
		
	\end{tabular}}
\vspace{-15px}
\end{table}

\noindent
{\bf Ablation Study.} We perform ablation study to verify the reliability of visual and language mutual guidance on each module. 
In the ablation study, we give the evaluation results of ground-truth setting on the RefCOCOg dataset. 
The results show in Table.~\ref{table:Ablation-study}. 
Line1 shows the result without mutual guidance. 
Line2$\sim$3 show the results of adding visual guidance and language guidance to subject module. 
The results show that visual guidance and language guidance all improve the comprehension of the model and prove the effectiveness of our method. 
Line4$\sim$5 shows the results of the same method applied to relation module and relationship module. 
We can see that the help for the improvement of model comprehension gradually decreases. 
The reason for this phenomenon is that of the three module weights generated by the language attention network, the subject module has the highest weight, the relationship module has the lowest weight and is less than 0.1 in most cases.

\label{sec:result}
\begin{figure*}[t]
	\centering
	\includegraphics[width=0.92\linewidth]{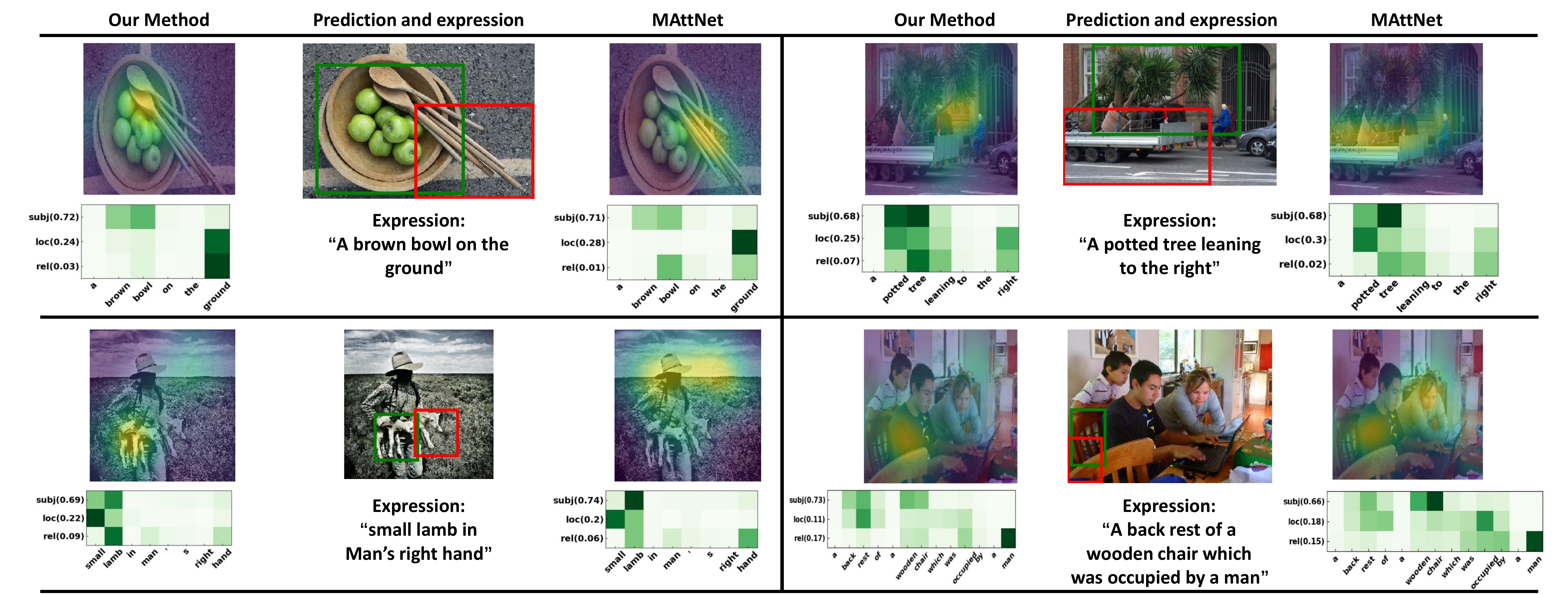} 
	\caption{Visualization comparisons between MutAtt and MAttNet of visual attention and language attention of each word on three modules. Green rectangle is the prediction result of MutAtt and red rectangle is the prediction result of MAttNet. We can see that MutAtt can adaptively capture the weight of each word, and accurately focus on the objects described in the language.}
	\label{fig:visualization}
	\vspace{-15px}
\end{figure*}


\subsection{Visualization}
We visualize the attention of image and the weights of expressions in Fig.~\ref{fig:visualization}. The first column is the comprehension result of our approach and the third column is the comprehension result of MAttNet. From the first set of examples, it is obvious that our method is superior to mattnet in terms of visual attention, language embedding and comprehensive understanding. With the guidance of ``a brown bowl on the ground", the focus area of the model is moved from the edge of the ``bowl" to the main body of the ``bowl". Correspondingly, with the help of the guidance of visual features, the model improves the understanding of the relationship between ``bowl" and ``ground" in ``a brown bowl on the ground", and encodes the ``ground" as a related object rather than a target object.



\section{Conclusion}
In this paper, we proposed a mutual attention-based guidance method (MutAtt) for the task of REC. 
MutAtt contains two key components for vision-language matching: \textit{visual-guided language embedding} and \textit{language-guided visual embedding}. By combining two matching processes, we maintains vision and language equality. So MutAtt can learn more discriminative visual feature and language embedding while guarantee vision-language consistency during the matching process in three sub-component, which beneficial to matching on cross-modal information. Experiments on three REC datasets with two setting show that MutAtt outperforms other method on most evaluation indicators, which demonstrates the effectiveness of MutAtt.

{
	\small
	\bibliographystyle{IEEEbib}
	\bibliography{Mutatt}
}
%


\end{document}